\documentclass{article}
\usepackage{ijcai19}

\usepackage{times}
\usepackage{soul}
\usepackage{url}
\usepackage[hidelinks]{hyperref}
\usepackage[utf8]{inputenc}
\usepackage[small]{caption}
\usepackage{graphicx}
\usepackage{amsmath}
\usepackage{booktabs}
\usepackage{algorithm}
\usepackage{multirow}

\usepackage{xcolor}
\usepackage{subfigure}

\usepackage{algorithmicx}
\usepackage{algpseudocode}
\usepackage{amsmath, amssymb}
\title{Hierarchical Meta Learning}% \\ for Generalizing Across Heterogeneous Tasks}

\author{
Yingtian Zou
\and
Jiashi Feng
\affiliations
National University of Singapore
\emails
\{elezouy,elefjia\}@nus.edu.sg
}

\begin{document}

\maketitle
\begin{abstract}
%We consider a new and challenging meta learning problem, \emph{i.e.}, how to obtain a meta model that can quickly learn to solve \emph{heterogeneous} tasks of novel structures. %Solving it is important for general artificial intelligence and also beneficial to many applications. However, 
Meta learning is a promising solution to few-shot learning problems. However, existing meta learning methods are  restricted to the scenarios where training and application tasks share the same output structure. To obtain a meta model applicable to the tasks with new structures, it is required to collect new training data and repeat the time-consuming meta training procedure.  This makes them inefficient or even inapplicable in learning to solve  heterogeneous few-shot learning tasks. 
We thus develop a novel and principled Hierarchical Meta Learning (HML) method. Different from  existing methods that only focus on optimizing the adaptability of a meta model to similar tasks, HML also explicitly optimizes its generalizability across heterogeneous tasks. To this end, HML first factorizes a set of similar training tasks into heterogeneous ones and trains the meta model over them at two levels to  maximize adaptation and generalization performance respectively. The resultant model can then directly generalize  to new tasks. %Moreover, to further alleviate the challenge from varying task structures, HML also learns a meta transformation over the internal representation from the meta model to facilitate adaptation to different task structures. 
Extensive experiments on few-shot classification and regression problems clearly demonstrate the superiority of HML over fine-tuning and state-of-the-art meta learning approaches in terms of generalization  across heterogeneous tasks. 
\end{abstract}

\section{Introduction}
Learning quickly to solve various tasks is a hallmark of general artificial intelligence, such as learning new skills from limited experience or learning to recognize new objects from a few examples. To achieve this, an artificial agent needs to effectively grasp generic and specific knowledge from  different tasks. 
As a promising solution, meta learning aims to learn a task-general meta model from multiple training tasks, serving as an inductive bias to improve the learning efficiency for new tasks~\cite{thrun2012learning}. Recently it has
received growing attention and achieved noticeable success in some machine learning applications, like model hyper-parameter optimization~\cite{maclaurin2015gradient,feurer2015initializing},  
reinforcement learning~\cite{duan2016rl,finn2017one} and few-shot learning~\cite{Sachin2017optimization,vinyals2016matching,finn2017model,santoro2016one}.  

Existing meta learning approaches mostly assume that the training and application tasks  share the same structure. The learned meta model therefore cannot be directly applied to learning to solve new  tasks of different structures.  
For example, a meta model trained for $N$-variable few-shot regression tasks is inapplicable to the new tasks of $N'$ variables. To obtain a new meta model that is compatible with the new task structure, the meta training process has to be repeated on tasks of new structures, which is highly inefficient. 
A desired efficient meta model should be able to solve  new tasks  with different structures quickly by learning broadly suitable prior knowledge in training phase, rather than confined to the tasks only with a familiar structure. Such a fundamental limitation hinders the application of meta learning in many realistic scenarios such as few-shot classification with a varying number of categories, or the task of learning a robot agent in a non-stationary environment where the  set of feasible actions continuously changes. However, to our best knowledge, such a problem of meta-learning for heterogeneous tasks has not been studied and solutions are still absent. 

In this work, we propose a novel meta learning approach, termed Hierarchical Meta Learning (HML), to address the above problem and generalize meta learning to heterogeneous tasks. 
Inspired by the state-of-the-art meta learning approach~\cite{finn2017model}, 
HML follows a new hierarchical meta learning scheme, in which a meta model is trained to have maximal performance on homogeneous new tasks, and meanwhile the generalizability is optimized over the tasks of novel structures. To make the generalization performance explicitly optimizable, HML constructs factorized task distributions from available  training tasks and trains the meta model on them hierarchically. %, as illustrated in Figure~\ref{fig_architecture}.  

Specifically, HML synchronously factorizes the tasks and training regime at two levels. At the bottom level, following~\cite{finn2017model}, HML optimizes sensitiveness of the meta model  $\theta$ to similar tasks for fast adaptation. At the top, HML takes as surrogate the performance of applying the model trained with one task distribution to the other different ones,  and explicitly maximizes the generalizability across heterogeneous tasks. In this way, the obtained meta model learns broadly applicable meta knowledge and can fast learn heterogeneous tasks of novel structures directly,  without repeating the expensive meta training. In addition, when applying to novel tasks with different structures, previous methods need to change the meta model architecture for compatibility. To alleviate such learning  difficulty, HML adopts a learnable transformation for internal representation of the meta model to enable fast adaption to new architectures.

%Our proposed HML is model agnostic. Here we particularly consider learning with neural network models due to their popularity. Meta learning for heterogeneous tasks thus faces another difficulty, \emph{i.e.}, the model architectures for heterogeneous tasks would be different and the meta model may suffer performance decline from   such architecture change. For instance, applying a meta model trained  for $N$-category classification to the $N'$-category one requires to replace the  classification layer with a new and compatible one. To alleviate such   difficulty, we introduce a meta   transformation layer on top of the meta model that is optimized for enabling fast adaption to new model architectures. 

To sum up, we make following  contributions:
\begin{itemize}
\item We are the first to propose the problem of meta learning for heterogeneous tasks. This problem features the critical limitation of existing meta learning approaches. Solutions to this problem would substantially extend application of meta learning. 

\item We propose a novel hierarchical meta learning approach to solve the problem. HML explicitly optimizes both the adaptability to homogeneous tasks and generalizability across heterogeneous tasks. Moreover, we introduce a meta  transformation function to further enhance the adaptability of the meta model to model architecture variation. 

\item We evaluate the HML with few-shot classification and regression experiments. Compared with state-of-the-arts, our HML significantly improves  generalization performance  to new tasks of different structures. 

\end{itemize}

\section{Related Work}

\paragraph{Meta Learning}

Recently, meta learning has drawn increasing attention, with which automatic learning schemes are devised to improve learning efficiency of existing learning methods or to learn (induce) the algorithms directly~\cite{pfahringer2000meta,lemke2015metalearning}. 
For example, 
\cite{finn2017model} targets at learning  how to initialize  a  model such that it can adapt to different tasks quickly through simple gradient descent fine-tuning. ~\cite{vinyals2016matching,snell2017prototypical} learn to match the query sample with the support ones based on metric learning in the embedded space.  Analogously,~\cite{rusu2018meta} tries to learn to initialize distribution parameters.~\cite{zhou2018deep} uses deep meta learning to learn the conception matching  of categories in the conception space. In our paper, we aim to obtain a meta model capable of generalizing across heterogeneous tasks.

\paragraph{Few Shot Learning}
The few shot learning (FSL) problem was first introduced by~\cite{fei2006one} and has received much attention since then.
%Despite its easy and natural properties in human intelligence, few-shot learning remains a challenging task for ``data hungry'' machine learning models because only a few training data points are available for training a complex model. 
To address this problem, some works~\cite{koch2015siamese,snell2017prototypical,garcia2017few,vinyals2016matching} develop models to learn discriminative  data representations and data matching rules. For instance, \cite{koch2015siamese} introduces a Siamese network to match the representation of support and query data which effectively circumvents the difficulty of learning a complex parametric classification model. \cite{vinyals2016matching} introduces the Matching Networks to learn to construct a support set to build a weighted nearest neighbor based classifier. As a follow-up, prototypical networks proposed by~\cite{snell2017prototypical} learn to embed data into a metric space where classification can be performed by associating the query with the nearest prototype representations of each class. Different from establishing a simple inductive bias in previous works, prototypical networks specifically model assumptions about the class conditional data distribution in the embedding space. Furthermore, some recent works  use the bidirectional LSTM~\cite{graves2005framewise} to embed the images and match the full context embedding by metric learning.  But all of them only target their models at the general few-shot classification setting and cannot be applied to regression problems. This work instead aims at developing a principled method that is applicable for both classification and regression.
%These methods prefer utilizing metric learning than defining the FSL as a regular classification problem. %Obviously, they focus 

\section{Hierarchical Meta Learning}

\subsection{Preliminaries}
% \JS{add a sub-section to introduce background of meta learning and MAML. }
% \JS{Define the notations to use here.}

%Let $ \mathcal{X} $ denote the input space and $ \mathcal{Y}$ be the label space. We are interested in meta learning for the $ N $-category  $k$-shot learning tasks, where only a small number of $k$ annotated  samples per category   (\emph{e.g.}, $k \leq 5$)  are available for  training a classification model within each task.

The goal of meta learning is to train a meta model that can quickly adapt to a new task using only a few samples and training iterations~\cite{pfahringer2000meta,lemke2015metalearning}. Formally, 
let $f_\theta(\cdot){:~} \mathcal{X} {\mapsto} \mathcal{Y}$ denote the  {meta}   model   with  parameter $ \theta $. Meta learning aims to optimize $\theta $ over a set of training tasks $T \sim p(T)$ with
\begin{equation}
\label{eqn:task_def}
 {T} \triangleq \{(x_{1},y_1),\ldots, (x_{n},y_{n}), (x_t,y_t), \ell\},
\end{equation} 
such that the  model $f_\theta$ is ready  to solve new tasks $T'=\{\{(x'_{1},y'_1),\ldots, (x'_{n},y'_{n}), \ell\}\} \sim p(T)$ by updating $\theta$ w.r.t $\{(x'_{1},y'_1),\ldots, (x'_{n},y'_{n})\}$ within a few steps.
Here $(x_j,y_j), j=1,\ldots,n$ are  task-specific training data, $(x_t,y_t)$ are reserved for training task evaluation and $\ell$ is the associated loss. An example of meta learning problems is the  $ N $-category  few-shot classification, where each task is to fast learn a classifier from a small number of $k$ annotated samples per category   (\emph{e.g.}, $k \leq 5$). 
%   More specifically, each task is to learn a  specific classification model   $ f_{\theta'}(\cdot) $ from only $ Nk $ training samples such that the following task-specific classification loss on  test samples $ ({x}_t, {y}_t) $ can be minimized:
% \begin{equation}
% \label{eqn:task_loss}
% \mathcal{L}_{{T}}(f_{\theta'}) \triangleq \ell(f_{\theta'}({x}_t),{y}_t),
% \end{equation}
% where $ \ell $ is the classification loss function.

% In meta training,  the meta model parameter  $ \theta$ is learned   to minimize the  \emph{meta loss} computed from all the training tasks:
% \begin{equation}
% \label{eqn:meta_loss}
% \theta  = \arg\min_\theta \sum_{i=1}^m  \mathcal{L}_{{T}_i}(f_{\theta_i'}) , {T}_i \sim p(\mathcal{T}),
% \end{equation}
% where $ \theta'_i $ is derived from the meta model $ \theta $ through task-specific adaptation,
% \emph{e.g.}, by  fine-tuning $ \theta $ on training samples of task $ {T}_i $. 

%We develop our Hierarchical Meta Learning (HML) algorithm from the state-of-the-art  MAML algorithm~\cite{finn2017model}, which is introduced here as preliminaries. 
%While we extend MAML in this work, our proposed idea is applicable to other meta-learning approaches.

As a state-of-the-art method, MAML~\cite{finn2017model} solves meta learning problems by optimizing the fast adaptability of the meta model $f_\theta$ such that it can solve a new  task rapidly via a few gradient descent steps on new tasks.  
To this end, in the meta model training phase, given a set of training tasks $ \mathcal{T} {=} \{ {T}_1, \ldots, {T}_m \} $, 
%where each task instantiates  an $N$-category $k$-shot  classification problem  $ {T}_i =\{(x^{(i)}_1,y^{(i)}_1),   \ldots, (x^{(i)}_{Nk},y^{(i)}_{Nk}), ({x}^{(i)}_{t},{y}^{(i)}_{t}), f_\theta, \ell \}$ as in Eqn.~\eqref{eqn:task_def}. 
 MAML fine-tunes the meta model  $f_\theta$  to a particular task ${T}_i$ by gradient descent at first:
\begin{equation}
  \label{eq:fine_tuning}
  \theta_i^\prime \leftarrow \theta - \alpha \nabla\mathcal{L}_{{T}_i}(f_\theta)
\end{equation}
where $\mathcal{L}_{{T}_i}(f_\theta) = \frac{1}{n} \sum_{j=1}^{n} \ell(f_\theta(x_j^{(i)}),y_j^{(i)})$ is the task-related training loss  and  $\alpha$ is a universal learning rate.

Then by treating each task as a training example, MAML optimizes $ \theta $ such that the following  meta loss for the  task-wise fine-tuned parameter $ \theta'_i $  over task-provided evaluation samples $(x_t,y_t)$ can be minimized:
% \begin{equation*}
% \min_\theta \sum_{i=1}^m \mathcal{L}_{{T}_i}(f_{\theta^\prime}) = \sum_{i=1}^m \mathcal{L}_{{T}_i}(f_{\theta - \alpha \nabla_\theta \mathcal{L}_{{T}_i}(f_\theta)}).
% \end{equation*}
\begin{equation*}
\min_\theta   \mathcal{L}_{MAML}(f_{\theta}) = \sum_{i=1}^m \ell(f_{\theta - \alpha \nabla_\theta \mathcal{L}_{{T}_i}(f_\theta)}(x_t^{(i)}),y_t^{(i)}).
\end{equation*}
%where $ \mathcal{L}_{{T}_i}(f_{\theta^\prime}) = \ell(f_{\theta^\prime}({x}^{(i)}_{t}), {y}^{(i)}_{t} ) $, \emph{i.e.}, classification loss on the reserved testing samples.
The meta parameter $\theta$ is then updated by gradient descent $\theta \gets \theta - \beta 
\nabla_\theta \mathcal{L}_{MAML}$.
 The  trained meta model $ f_\theta $    can be applied directly to a new \emph{similar} meta learning  tasks through gradient descent  fine-tuning  in Eqn.~\eqref{eq:fine_tuning}.
 %and performs remarkably well.

\subsection{Problem Setting}
Existing meta learning approaches require the training tasks $T$ and new tasks $T'$ to be from the same distribution $p(T)$ and share  \emph{the same structure}. Such restriction makes existing approaches, such as MAML, inapplicable to novel tasks of different structures. %For instance, in few-shot meta learning,  the model can only solve classification problems with the same number of categories.  

To overcome such a limitation of meta learning,  we introduce  a  more general problem where the meta model will be tested on tasks with different structures from the training ones. This problem is very common in practice. For example, one may wish to construct a meta model for $N'$-category few-shot classification but the training data are only sufficient for constructing several $N$-category   classification tasks with $N \ll N'$; or one needs to build a navigation agent with only very few feasible actions in the training environments but faced with testing environments that are more complex and allow more actions.  
%In addition to more closely fitting the realistic application scenarios, solutions to this problem setting would significantly broaden application scope and enhance efficiency of meta learning. 
Solving these problems requires a novel meta learning scheme with stronger generalization guarantees\textemdash although the  meta model only has access to  tasks $T\sim p(T)$ for training but it can efficiently adapt to  heterogeneous tasks $T'\sim p'(T)$  of different structures. %, even the model is not  meta-trained on a collection of tasks $T'$.
%For instance, in few-shot learning,   we target at a meta model that is trained on  $N$-category  classification tasks but tested on   $N'$-category classification tasks with $N'\neq N$.  
%We aim to obtain a meta model $f_\theta$ which is trained only on tasks $T \sim p_i(T)$ and can be applied directly to solve tasks  $T' \sim p_j(T)$ quickly. 
Formally, with only similar tasks $T\sim p(T)$ available, we aim to obtain a meta model $\theta$ that can adapt to different tasks $T' \sim p'(T)$ efficiently through a few gradient-descent steps and minimize the  generalization loss: 
\begin{equation}
\label{eqn:gen_loss_intro}
\min  \mathcal{L}_{gen} \triangleq   \sum_{T'\sim p'(T)} \ell(f_{\theta - \alpha \nabla_\theta \mathcal{L}_{{T}'}(f_\theta)}(x'_t),y'_t).
%\mathcal{L}_{T'}(f_\theta) 
\end{equation}
This problem is quite challenging since the test tasks  $T'$ are not available in advance. In the following, we develop a new hierarchical  meta learning approach to address it. 
%The above loss measures the generalization performance of the meta model $\theta$. 
%where $\varphi_j$ is the classifier head that is compatible with the task structure (\emph{i.e.}, the number of categories in  the few-shot classification setting).

%To concern about the development of few-shot learning, we hope to propose a new meta learning rule to address issue of multi types tasks. 

\subsection{Hierarchical Meta Learning}
%\JS{I think the general rule in your mind is similar to ``learning to learn'' in meta learning. }
%A core characteristic of meta learning is to  learn a general rule for solving various tasks~\cite{pfahringer2000meta,lemke2015metalearning}. Existing meta learning approaches  usually target at learning the rule that is applicable to a type of tasks with the same structure~\cite{vinyals2016matching,snell2017prototypical,finn2017model}. Thus they would fail if the new tasks present different structures from the training ones.

To address the above problem, we develop a novel hierarchical meta learning (HML) approach to  learn a rule that is generalizable across heterogeneous tasks and meanwhile is suitable for similar tasks.  HML achieves this  by jointly optimizing the adaption performance to homogeneous tasks and the generalizability across heterogeneous tasks of the meta model, as depicted in the following learning objective:
\begin{equation*}
\min_\theta \sum_{T \sim p(T)} \ell(f_{\theta - \alpha \nabla_\theta \mathcal{L}_{{T}}(f_\theta)}(x_t),y_t)  + \sum_{T' \sim p'(T)} \mathcal{L}_{\theta_{T}\rightarrow T' }.
\end{equation*}
%where the second term measures how well the meta model $f_{\theta_{T}}$ trained on tasks $T$ generalizes to new tasks $T'$ with different structure. 
Here, the second term is the generalization loss defined in Eqn.~\eqref{eqn:gen_loss_intro}:
\begin{equation}
\label{eqn:gen_loss}
\mathcal{L}_{\theta_{T}\rightarrow T' } =  \sum_{T'\sim p'(T)} \ell(f_{\theta_{T} - \alpha \nabla_{\theta_{T}} \mathcal{L}_{{T}'}(f_{\theta_{T}})}(x'_t),y'_t).
\end{equation}
% \begin{equation*}
% \begin{aligned}
% \min_\theta \mathcal{L}_{HML} &= \sum_{T_i \sim p_i(T)} \ell(f_{\theta - \alpha \nabla_\theta \mathcal{L}_{{T}_i}(f_\theta)}(x_t^{(i)}),y_t^{(i)}) \\
%  &\qquad + \sum_{j\neq i} \sum_{T'\sim p_j(T)} \ell(f_{\theta - \alpha \nabla_\theta \mathcal{L}_{{T}'}(f_\theta)}(x'_t),y'_t).
% \end{aligned}
% \end{equation*}
The challenge of optimizing the above objective lies in the fact that the tasks $T'\sim p'(T)$ are not available in the meta model training phase, making the generalization loss not directly optimizable. To alleviate this challenge, we propose to sample a set of training tasks, denoted as $\mathcal{T}$,  from the available distribution $T \sim p(T)$ and restructure these tasks into two-level hierarchies $\widehat{\mathcal{T}} = \{\mathcal{T}_1, \mathcal{T}_2 , \ldots, \mathcal{T}_H \}$ where $\mathcal{T}_h = \{T^{h}_1, \ldots, T^{h}_m\}$ for $h=1,\ldots, H$. Here the top-level tasks $\mathcal{T}_1,\ldots, \mathcal{T}_H$ have different structures and each of them consists of homogeneous tasks, \emph{i.e.}, $T^{h}_1, \ldots, T^{h}_m$ have the same structure.  In particular, we choose to factorize the output space to construct the heterogeneous top-level tasks. Recall each task $T\in \mathcal{T}$ is to learn a mapping function $f_\theta: \mathcal{X} \mapsto \mathcal{Y}$. We factorize $\mathcal{Y} $ into $\mathcal{Y}_1, \ldots, \mathcal{Y}_H$ and require $\mathcal{Y}_1 \subset \mathcal{Y}_2 \subset \ldots \subset \mathcal{Y}_H=\mathcal{Y}$ w.l.o.g. Then the task $\mathcal{T}_h$ corresponds to learning a mapping function $f_{\theta}: \mathcal{X} \mapsto \mathcal{Y}_h$, and $\mathcal{T}_H = \mathcal{T}$. Take the $N$-category few-shot learning problem as an example. Each task is to learn to solve an $N$-category classification problem. One thus can extract heterogeneous tasks of  2-category, 3-category to $N$-category few-shot learning from the available tasks to form the task hierarchy.

%The top level lies heterogeneous tasks with different structures and the bottom level presents homogeneous tasks. 
%Homogeneous tasks with different categories is ``subtasks" and heterogeneous tasks with different classes number named ``tasks". A set of homogeneous subtasks warped to the tasks. 
Instead of optimizing the meta model $f_\theta$ on tasks $\mathcal{T}$ directly as conventional meta learning approaches, HML performs meta learning on the hierarchy of re-structured tasks that demonstrate task heterogeneity.  % An illustration of such a task hierarchy is given in Figure~\ref{fig_architecture}. 
The purpose of extracting and aggregating tasks in this way is  to facilitate the hierarchical meta learning, which offers heterogeneous tasks systematically and makes optimization of the generalization meta objective feasible. The meta learning objective of HML becomes
\begin{equation*}
\begin{aligned}
& \min_\theta \mathcal{L}_{HML} \triangleq \\
& \sum_{T \in \mathcal{T} } \ell(f_{\theta - \alpha \nabla_\theta \mathcal{L}_{{T}}(f_\theta)}(x_t),y_t) + \sum_{h=1}^{H-1}  \sum_{T^h \in \mathcal{T}_h } \mathcal{L}_{\theta_{T^h} \rightarrow T^{h+1}}.
\end{aligned}
\end{equation*}
%As it shown in Fig.\textcolor{red}{~\ref{fig_architecture}}, there are many adaptive parameters showing the multilevel optimization routes. 
The loss has two components. The first one requires the meta model $\theta$ to fast adapt to the training tasks quickly, which is similar to MAML. The second one requires the meta model trained on tasks ${T}^{h-1}$ to be able to quickly solve new tasks ${T}^{h}$ with a different structure. Through minimizing this hierarchical loss, the meta model is exposed to multiple meta learning tasks and more importantly, the cross-task learning scenarios. 
During meta training phase, HML is trying to get the optimal solution for every low-level task $T^h$. In this way, the low-level meta model learns the generalized solution on the task set. High-level meta learner asynchronously learns the global generalized solution over all the heterogeneous tasks. 
%Besides, we bridge the different task domain by inserting an intermediate layer to better comprehend the transfer rule. 
Thus, besides learning the rule for few-shot classification, the HML model also learns the rule of generalizing to heterogeneous tasks. The resulted  meta model gains the ability to solve homogeneous and heterogeneous tasks. 

For optimization, we first sample a batch of low-level tasks $\{T_i, i{=}1,\ldots,m\}$. 
% Given the representation learner $\theta^n$ (adapted to the task $n$) and corresponding classifier $\varphi^n$, the prediction will be $f_{\theta^n,\varphi^n}$. To efficiently learn the meta model to solve few-shot classification tasks, we draw the inspiration from MAML~\cite{finn2017model} to meta learn the initial model parameters $\theta$. Specifically, 
Then, we compute  meta gradients $ \nabla \mathcal{L}_{HML, T_i}$  and update the meta model parameter as
%Then the initialized parameter of learner can be meta learned through steering vector sum of meta gradient. In this way, we can get parameterized formula:
\begin{equation}
\label{eq_theta}
\theta \gets \theta - \beta \sum^m_{i=1} \nabla_{\theta} \nabla \mathcal{L}_{HML, T_i},
\end{equation}
where $\beta $ is the meta learning rate. 

\paragraph{Meta Model Transformation}

When applied to another task $T' \sim p'(T)$ with a different structure, the meta model  $\theta$ from vanilla meta learning is not directly applicable, because $\theta$ cannot adapt to such unseen tasks. But with HML, the meta model parameter $\theta$ will converge to a more generalized position in the representation space which easily transfers to various task distribution and adapts to new tasks quickly. 
%Given a training batch of high-level tasks \{$T^n, n=1,...,N$\}, we can initial equal amounts of classifiers $\{\varphi^n,n=1,...,N\}$. %写通用
%In general, the global optimization objective is suggested for this purpose:
%$$\mathcal{L}_{\sum T^n} = \sum\limits^N_{n=1} f(\theta^*,\varphi^n,T^n) $$

%\subsection{Learning Generalization Rule}
% As  mentioned above, how to learn a rule is the core idea of meta learning. HML devises a generalization rule which guides the meta model to obtain the transfer ability across different task structures. %引导学习者拥有在不同task domain之间迁移的能力
% In essence, HML learns  two types of generalization: 
% \begin{itemize}
% \item Hierarchical learning  a high-level adaptive parameters which can be optimized to adapt  to various task distributions  in a few  steps.
% \item A transform layer bridges two different task distributions. Thus this layer tries to learn a solution of generalizing the representation.
% \end{itemize}
% For the second point, we wish to explicitly develop the generalization ability of the model. But the structure of the applied  tasks is  unavailable during training phase or may constantly vary. Thus HML meta learns from  known task distributions,   trying to learn generalization rule across heterogeneous  tasks. 
%In order to learn a transferable representation, we construct cross domain task to generalize it.  Specifically, we construct cross domain task like $T^n\to T^{n+c}, n+c\leq N$, denoted as $\hat{T}^{n(c)}$. Suppose we get the representation $f(\theta,T^n)$, and it will be fed to a classifier $\varphi^n$ in former task $T^n$. 

HML gives the optimal meta model parameter $\theta$ that can adapt to different task distributions. When applying to a new task, the architecture of the output needs to change from the trained tasks due to the task structure difference. This brings another challenge for learning to generalize across heterogeneous tasks.
Indeed, generalizing across tasks means the meta model $\theta$ is trained on the training task $T$ at first. Then it will adapt to different tasks $T'$. It is not wise to disorder the parameters by training the meta model on heterogeneous tasks directly as this may bring the noisy task obstructing. To address this difficulty, we explicitly separate meta model parameters into $\theta $ and $\varphi$ where $\varphi$ connects to the output directly (\emph{e.g.}, the output layer of a neural network model).  We propose to add  an intermediate \emph{transformation} function $\omega$ between the meta model $\theta $ and the output function  $\varphi$   to transfer   smoothly to different tasks\footnote{When applying to neural network models, we use a convolution layer and  linear activation  units~\cite{glorot2011deep} to implement the transformation function $\omega$ that does not change the size of input feature maps.}. 
  %In application,  $\varphi$ which is usually randomly initialized. To fast adapt to the change of the output structure, we introduce another transformation function  $\omega$ on top of $\theta$. 
  %This function is meta learned to adapt to different $\varphi$ quickly.

Suppose we have the tasks $\mathcal{T}_1, \ldots, \mathcal{T}_H$ constructed as above and let $\varphi_h$ denote the output function compatible with tasks $T \in \mathcal{T}_h$.  The transformation function $\omega$ is trained to optimize the following objective. Firstly, the meta model $\theta$ is adapted to a specific task $T^h $ by 
%transfered to new representation $f(\theta^*,\Omega,\hat{T}^n)$ over new task $\hat{T}^n$. To better mapping the output feature, transform layer should map it to more complex task $T$. 
%Besides, random initialized classifier means it is hard to find the appropriate solution for fast adaptation in new task domain. When we transfer  there would be a new error from the untrained classifier. There, we only need to fine-tune the transform layer rather adding extra tasks on $\theta^*$. Based on this hypothesis, we apply the similar meta learning way on transform layer. Firstly, for each task $\hat{T}^n, n=1,...,N$ the meta loss is:
\begin{equation}
\label{eq_}
\begin{aligned}
\theta^h \gets \theta - \nabla_\theta \mathcal{L}_{T^h}( f_{\theta,\varphi^h}). 
\end{aligned}
\end{equation}
Secondly, the transformation function $\omega$ should map the meta model $\theta $ to a different task $T^{h+1}$ as follows:
\begin{equation}
\label{eq_}
\begin{aligned}
\theta^{h+1} \gets \theta^h - \nabla_\theta \mathcal{L}_{T^{h+1}} f(\omega(\theta^h),  \varphi^{h+1}).
\end{aligned}
\end{equation}
Thus the loss for optimizing $\omega$ is computed as
\begin{equation}
\label{eqn:omega_loss}
\begin{aligned}
\mathcal{L}_{\omega} = \sum_{h} \ell({f_{\omega(\theta^{h+1}), \varphi^{h+1}}(x_t^{h+1}),y_t^{h+1}  )}.
\end{aligned}
\end{equation}
We optimize $\omega$ through gradient descent.
To enable the transformation function with parameter $\omega$ to fully learn the generalization rule, we just update the parameter of $\omega$  while fixing other parameters $\theta$ and  $\varphi^h$. 
% \begin{equation}
% \label{eq_}
% \begin{aligned}
% \omega &\gets \omega - \nabla_\omega \mathcal{L}_{\omega}.
% % \Omega &\gets \Omega - \gamma\sum\limits^N_{n=1} \nabla_\Omega \mathcal{L}_{\sum \hat{T}^n} \\
% % \Omega^n_i &\gets \Omega^n_i - \gamma_i\sum\limits^N_{n=1} f(\theta^*,\Omega,\hat{\varphi}^n,\hat{T}^n) 
% \end{aligned}
% \end{equation}
%In a nutshell, we proposed a generalization rule that hope it can help model generalize to heterogeneous task by meta learning way. This rule utilizes known tasks domain to perform the transfer learning in training. But to circumvent the disorder in optimization, we introduce the transform layer $\Omega$ to serve as the generalization leaner.

%the parameter of classifier is incompatible with feature extractor. we can not specify the category number. . Thus , . 

\paragraph{Training Algorithms}
% Hierarchical learning and generalization rule seem to be an intractable problem. 
HML optimizes the meta model $\theta$ and transformation function $\omega$ alternatively. The details are summarized in Algorithm \ref{alg:HML}.% and Algorithm \ref{alg:omega} respectively. 

\renewcommand{\algorithmicrequire}{\textbf{Input:}}
\renewcommand{\algorithmicensure}{\textbf{Output:}}
\begin{algorithm}[h]
	\caption{Hierarchical Meta Learning }
    \label{alg:HML}
	\begin{algorithmic}[1] %每行显示行号
	\Require Task distribution $p(T)$, meta training batch size $M$, number of sub-tasks $H$.
    \Require step size $\alpha, \beta, \gamma$
    %Model parameter $\theta$, $\varphi_n(n = 1,...,N)$, generalizing layer parameter $\Omega$, meta batch size $M$, tasks samples $T^{(m)}_n(n=2,...,N, m=1,...,M) \in \mathcal{T}(n)$ 
	\Ensure Meta model parameter $\theta^*$
	\State Randomly initialize $\theta, \varphi^h, h=1,\ldots, H$ %\Call{Gaussian Initialize}{ $\theta$, $\omega$, $\varphi$ }
	\While{not done}
		%\For{$n = 1, ... ,N$}
        	\For {$m = 1,\ldots,M$}
				\State Randomly sample  task $T_m \sim p(T)$
                \State Obtain sub-tasks $T_m^1, \ldots, T_m^H$
                \For {$h = 1,\ldots,H-1$}
                \State Adapt meta model $\theta$ to task $T^{h}_m$ by: 
                %\State $G_{\theta^m} =\nabla_{\theta} \mathcal{L}_{a}f(\theta^m,\varphi^m_n,T^{(n)}_m)$
                \State $(\theta^h,\varphi^h) \gets (\theta,\varphi^h) -\alpha  \nabla \mathcal{L}_{T^h_m}f_{\theta,\varphi}$
                \State Evaluate $\mathcal{L}_{\theta^h \rightarrow T^{h+1}}$ in Eqn.~\eqref{eqn:gen_loss}.
\EndFor
 \State Adapt meta model $\theta$ to task $T^{H}_m = T_m $ by
  \State $(\theta^H,\varphi^H) \gets (\theta,\varphi^H) - \alpha \nabla  \mathcal{L}_{T^H_m}f_{\theta,\varphi}$
  \State Evaluate $\ell_{T_m}(f_{\theta^H, \varphi^H}(x_t), y_t)$
  \State Evaluate loss $\mathcal{L}_\omega $ in Eqn.~\eqref{eqn:omega_loss}.
                \State Update $\omega \gets \omega - \gamma \nabla \mathcal{L}_\omega$
            \EndFor
            \State Compute meta loss:
            \State $\mathcal{L}_{HML} {=} \sum_{m}\ell_{T_m}(f_{\theta^H, \varphi^H}(x_t), y_t) {+} \sum_h \mathcal{L}_{\theta^h \rightarrow T^{h+1}}$
            \State Update $\theta \gets \theta - \beta \nabla_\theta \mathcal{L}_{HML}$
            \State Update $ \varphi^h \gets \varphi^h - \beta \nabla_{\varphi^h}  \ell_{T_m}(f_{\theta^h, \varphi^h}(x_t), y_t)$
	\EndWhile
	\end{algorithmic}
\end{algorithm}

\section{Experiments}
We first investigate whether the proposed HML can generalize well to new tasks with different structures from the training ones through experiments with few-shot classification on three benchmark datasets and few-shot regression. We also evaluate the effectiveness of the proposed  hierarchical meta learning and meta model transformation. Finally, we present experiments to explain how the meta model from HML obtains the ability of generalization across heterogeneous tasks.

\subsection{Datasets} 
%We evaluate our model on the standard $N$-way $k$-shot meta learning tasks~\cite{vinyals2016matching}. Each task is to learn an $N$-category classification model from only $Nk$ training examples, where $k$ is the number of training examples  per category and typically is very small (\emph{e.g.}, 1 or 5).

We use three datasets for few-shot classification evaluation. The Omniglot dataset~\cite{lake2011one} consists of 1,623 characters (categories) from 50 different alphabets and each character is drawn by 20 different subjects. Following~\cite{lake2015human}, we form the training set with 30 alphabets and use the rest alphabets for testing. 
%For $N$-way $k$-shot task, $N$ is the number of characters. 
The second dataset is the miniImageNet~\cite{vinyals2016matching} which  consists of 60,000 color images of size 84$\times $84 from 100 categories. Each category has  600 examples. In our experiment, we  split the 100 categories into 64, 12 and 24 randomly. The first 64 categories are used for making training tasks, and the 12 categories are used for validation and 24 for testing. The third dataset is the SUN2012~\cite{xiao2010sun}  containing 899
categories and 130,519 images in total. To construct few-shot learning tasks, we remove the categories whose example images are less than 6. We obtain  441 training classes and 65 testing classes which contain 11,226 images and 613 images respectively. We randomly select categories from the training set and test set to form the training and test learning tasks respectively. 
%Note that training set has access to all images under training classes. $k$ images in each classes of testing set is the fine-tune set.

\subsection{Evaluation Protocol}
%Following the standard $N$-way $k$-shot setting~\cite{vinyals2016matching}, every task contains $N\times k$ training images with $k$ images per class. 
Under the conventional few-shot meta learning evaluation setting, the training tasks share the same structure as the test tasks. More specifically,  a meta model is trained on a collection of $N$-way $k$-shot learning tasks from the training set and then evaluated on multiple $N$-way $k$-shot   tasks sampled from the test set. Here $N$ is fixed for training and test.

%As for high-level tasks, different task is $n$-way $k$-shot where $n=1,...,N$.  as its 
In this work, we aim to develop a meta model capable of fast solving new tasks of different structures. To evaluate its generalization performance, we construct the following evaluation protocol. During the meta model training phase, similar to the conventional setting, a collection of  $N$-way $k$-shot learning tasks from the training set are provided. Across all the experiments, we set $N=5$.
%For the sake of evaluation, performance metric restrict different tasks have different number of classes but same examples per class. All the examples in training set is used to learn the representation. 
During testing, we randomly sample $N'$-way $k$-shot tasks from the test set. Here the values of $N'$ are varied and different from $N$ of the training tasks. We are  particularly interested in the model generalizability from simpler classification tasks to more complex ones. Therefore, we set $N' \geq N$ across all the evaluations. In particular, we adopt $N'=5, 10, 20$ and $50$ on the Omniglot dataset. On the miniImageNet and SUN datasets, we evaluate the models for $N'=5, 6, \ldots, 10$. We report the  classification averaged accuracy for each type of  test tasks over $10$ random train/test splits. 

% images without replacement from every test class to fine-tune or adapt the models. And the accuracy performance gets results from average accuracy of multi evaluation.
%For HML, we sampled few shot learning task uniformly at random

%Initialized classifier was trained on the tasks whose categories is outnumber  

\subsection{Baselines and Model Architecture}
To our best knowledge, there is no  solution ready for coping with heterogeneous tasks without requiring to repeat meta training. For evaluation and comparison, we construct following competitive baselines. The first  is fine-tuning. The model is trained on the meta-training tasks and then fine-tuned on heterogeneous test tasks. The second one is MAML~\cite{finn2017model} which has been shown to excel at adapting to new tasks swiftly across a variety of meta learning problems. We use MAML to obtain a meta model and fine-tune it on the test tasks by a few steps of gradient descent.  We replicate MAML following source code\footnote{\url{https://github.com/cbfinn/maml}} and achieve the results reported in~\cite{finn2017model}. The third is Meta-SGD~\cite{li2017meta}.
%from the authors to implement MAML and all the hyperparameters for the baselines are tuned to be optimal.
%Even so, they didn't address this problem with an approach akin to generalize across heterogeneous tasks. In our experiments, MAML is trained on different $5$-way $k$-shot learning tasks. The well-trained parameter was then adapted to new classifiers to solve $5,10,20$-way newly tasks. 

%Our HML was trained on heterogeneous tasks ranging categories from $2-5$. As it mentioned above, we trained generalizing layer by calculating meta loss on more complex tasks. After that, we adapt it to $5,10,20$-way newly tasks and test it respectively.

We use a deep CNN architecture as backbone  to implement different meta learning approaches. The network consists of three convolution layers with pooling operations and one fully-connected layer. 
% We trained this net on random selected 5 categories with all training images and fine-tuned it on new tasks with only $n\times k$ images so named it ``Fine-tune''. Second baseline is MAML~\cite{finn2017model}, we follow the paper trained it on $5$-way tasks. And then it will adapt to new tasks with $k$ image per class. 
For HML, we use a convolution layer to implement the meta transformation function $\omega$, with stride  1 and kernel size  1$\times $1. When applying to novel test tasks, we  replace the trained classifier (\emph{i.e.}, the output layer) with new task-compatible classifiers.

\subsection{Few-shot Classification}
% To completely evaluate our model performance, we evaluate it over three datasets. And we also establish three competitive model as the state-of-the-art. Here is the results of different experiments of different datasets to demonstrate our model's performance relative to the state-of-the-arts.
To verify the effectiveness of our HML on obtaining well-generalizable meta models across heterogeneous tasks, we first compare with strong baselines over the few shot classification problem.

\subsubsection{Results on Omniglot}
The results on Omniglot are given in Table~\ref{table_comparison_omni}. As  observed, meta learning methods show significant superiority to the fine-tuning, benefiting from the explicitly optimized adaptability to new tasks. However, MAML and Meta-SGD perform not well when applying to new tasks with different structures (\emph{i.e.}, different $N'$). In particular, the meta model trained on $1$-shot $5$-way tasks performs poorly on $20$-way and $50$-way tasks, manifesting such heterogeneity of task structures raises significant challenges to state-of-the-art meta learning approaches. In contrast, our HML performs much better than other meta models, even though all of the models are trained on $5$-way tasks.  
%From $5$ to $50$-way tasks we can see, deterioration of MAML's accuracy performance is more serious than our method. 
For the  $20$-way and $50$-way $1$-shot learning tasks,  our HML outperforms MAML by a large margin of 5.5\% and 14.4\%  respectively. We have similar observations for the $5$-shot experiments. HML also outperforms the state-of-the-art by 3.6\% and 12.9\% on $20$-way and $50$-way  learning tasks. Based on the observation that the gap grows notably as the classes number increases, it can be seen that the meta model obtained from HML has a stronger generalizability, and also that our method is indeed effective at lifting the meta models to solve heterogeneous tasks.

\begin{table}[h]
\caption{Few-shot classification accuracy (\%) on  Omniglot. The models are trained on 5-way tasks and evaluated on  tasks ranging from 5-way to 50-way.}
\small
\begin{tabular*}{0.46\textwidth}{@{\extracolsep{\fill}}ccccc}
\toprule
1-shot & 5-way & 10-way & 20-way & 50-way \\ \midrule
Fine-tune & 62 & 57 & 29.5 & 9.2 \\
Meta-SGD & \textbf{99.3} & 85 & 69.5 & 52.7 \\
MAML & 98 & 90 & 79.5 & 56.8 \\
HML (ours) & 98 & \textbf{92.5} & \textbf{85} & \textbf{71.2} \\

 \bottomrule
 \toprule
5-shot & 5-way & 10-way & 20-way & 50-way \\ \midrule
Fine-tune & 81.2 & 70.6 & 44.3 & 20.2 \\
Meta-SGD & \textbf{99.8} & 91.6 & 85.3 & 66.5 \\
MAML & 99 & 92.4 & 86.6 & 68.2 \\
HML (ours) & 98.3 & \textbf{94.6} & \textbf{90.2} & \textbf{81.1} \\ \bottomrule
\end{tabular*}%
\label{table_comparison_omni}
\end{table}

\subsubsection{Results on miniImageNet}
Few-shot learning on miniImageNet is a more challenging task than that on Omniglot  as  images from miniImageNet present much more complex contents. We train the meta models on the $5$-way tasks and evaluate them on $6,7,8,9,10$-way tasks with greater evaluation granularity (only $5,10-way$ reported due to the limited space). As shown in the top panel of Table~\ref{table_comparison_mini}, the performance of the fine-tuned model rapidly falls below  $10 \%$ as task difference becomes larger. In contrast, MAML shows superiority as it has fast adaptability: MAML is $19.2\%$ higher than the fine-tuned model when extending to $10$-way task. Even so, MAML still suffers inferior performance due to its lack of generalization power across new task structures. MAML has a conspicuous decline from $43.6\%$ to $22.2\%$ and $62.9\%$ to $37.5\%$ on $1,5$-shot respectively. Its degradation on $5$-shot ($25.4\%$) is more severe than the fine-tuned model ($14.8\%$). Therefore, we can say HML offers convincing performance enhancement w.r.t. MAML and the fine-tuned model.

\begin{table}[h]
\caption{Few-shot classification accuracy (\%) on  MiniImageNet and SUN2012. The models are trained on 5-way tasks and evaluated on  tasks ranging from 5-way to 10-way. 10-way 1-shot task is abbreviated as ``10w1s".}
\small
\begin{tabular*}{0.46\textwidth}{@{\extracolsep{\fill}}ccccc}
\toprule
MiniImageNet & \small 5w1s         & 10w1s         &\small  5w5s          & 10w5s \\ \midrule
Fine-tune   & 29.6          & 7.0            & 28.6          & 9.2 \\
MAML        & 46.3            & 22.2            & 62.9        & 37.5 \\
Meta-SGD        & \textbf{47.6}            & 22.2            & \textbf{64.01}        & 32.4 \\
HML (ours)  & 46.6   & \textbf{28.8} & 60.4 & \textbf{41.8}  \\
\midrule
SUN2012     &\small 5w1s & 10w1s & \small 5w5s &  10w5s \\ \midrule
Fine-tune   & 32  & 14.0 & 41.2 & 24.2 \\
MAML        & 49.9 & 32.5 & 64.6  & 39.8 \\
Meta-SGD        & \textbf{51.3}            & 32.9            & \textbf{66.8}        & 37.7 \\
HML (ours)  & 48.6  & \textbf{39.5}  & 65.4 & \textbf{45.2} \\
\bottomrule
\end{tabular*}%
\label{table_comparison_mini}
\end{table}
\subsubsection{Results on SUN2012}
%In this paper, we adopt a part of SUN2012 dataset as few-shot learning dataset by removing the classes contains images less than 10. In this way, we randomly split dataset as  training set (contains 441 classes) and testing set (contains 65 classes). 
Similar to miniImageNet, we also set $5,10$-way tasks as the incremental classifier learning tasks. The results of $N'$-way $1,5$-shot tasks are given in Table~\ref{table_comparison_mini}. These results provide consistent evidence that  HML is effective and provides a better solution to obtaining meta models that can generalize across heterogeneous tasks.
%On the other hand, there is still room for the performance on new tasks to improve. This demonstrates that generalization across heterogeneous tasks is a challenging problem for meta learning and more research efforts are needed. 
%HML has a fast adaptation ability with newly initialized classifiers. As for setting, 
Our new few-shot learning setting can serve as a  standard testbed to evaluate the generalizability of meta models, which better aligns with realistic application scenarios.

% \begin{figure*}[t!]
%     \centering
%     \begin{subfigure}[t]{0.5\textwidth}
%         \centering
%         \includegraphics[height=1.2in]{maml.png}
%         \caption{20-way new tasks t-SNE of MAML}
%     \end{subfigure}%
%     ~ 
%     \begin{subfigure}[t]{0.5\textwidth}
%         \centering
%         \includegraphics[height=1.2in]{40plus.png}
%         \caption{20-way new tasks t-SNE of HML}
%     \end{subfigure}
%     \caption{20-way new tasks t-SNE}
% \end{figure*}

\begin{figure}[t]
\centering
\subfigure[MAML]{
\includegraphics[width=0.25\textwidth]{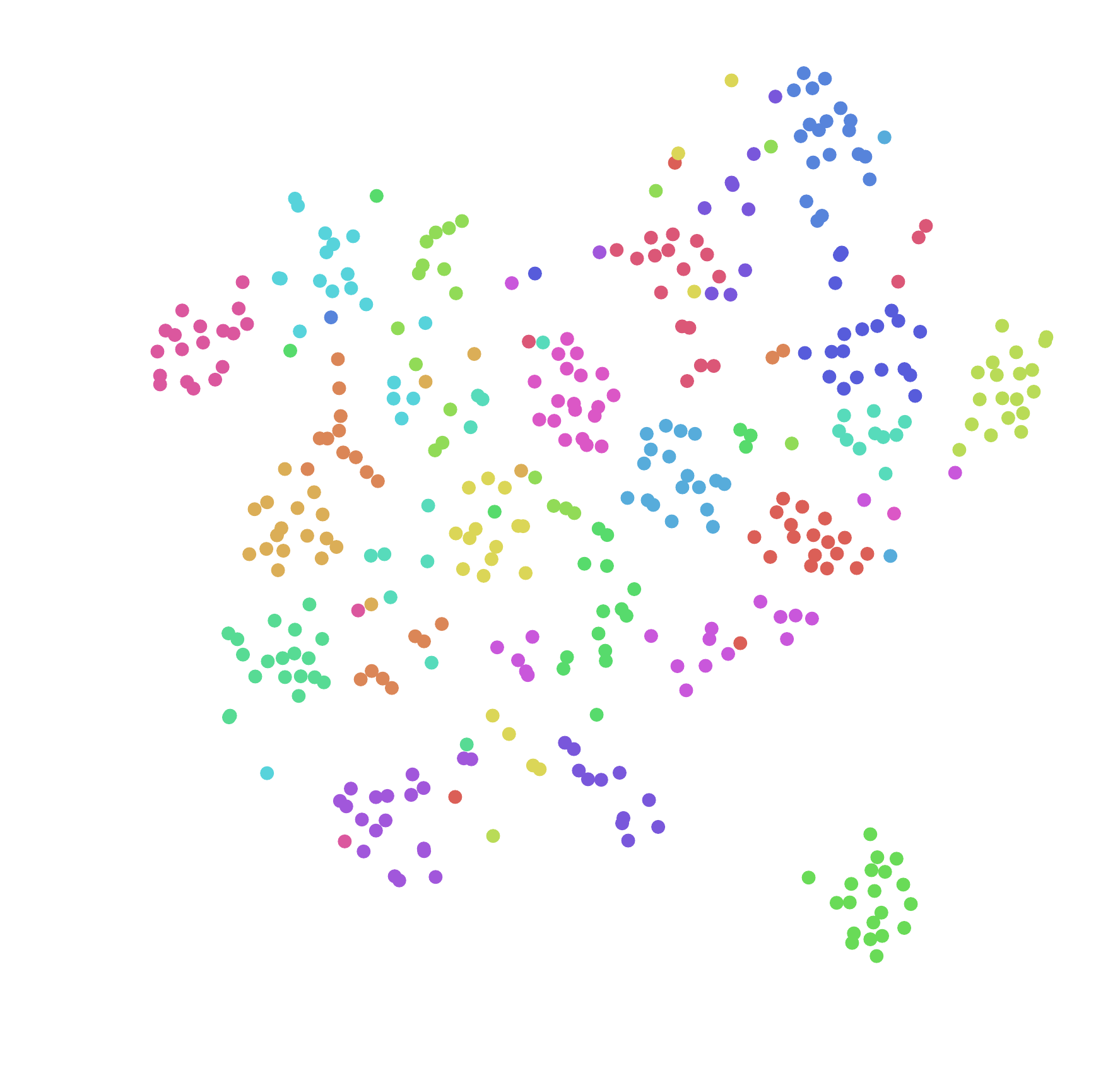}
}
\hspace{-10mm}
% \caption{20-way new tasks t-SNE of MAML}
\subfigure[HML]{
\includegraphics[width=0.25\textwidth]{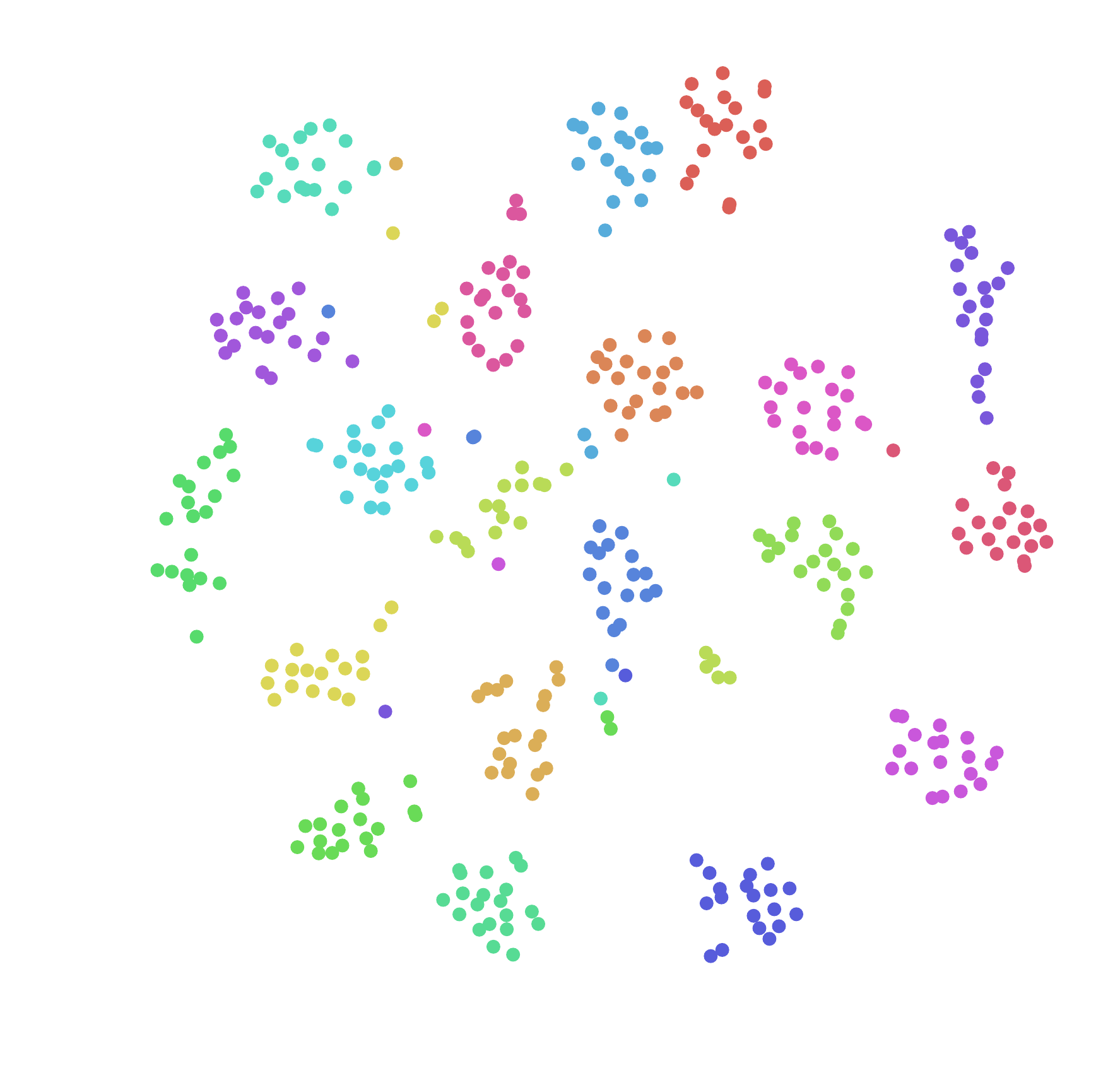}
}
\caption{The t-SNE on the data representations provided by the meta model for $20$-way $1$-shot tasks. The meta model is trained using MAML and HML on $5$-way $1$-shot tasks. }
\label{fig_result_maml}
\end{figure}

\subsection{Few-shot Regression}
We further evaluate  our proposed HML on multivariate linear regression tasks,  to demonstrate the meta model obtained from HML is also efficient in learning new tasks for regression problems.  Specifically, given 5  independent variables $X \in \mathbb{R}^{d_x }$ of dimensionality $d_x$ and corresponding output $Y\in \mathbb{R}^{d_y}$, the task is to fit the linear function $Y = w^\top X + b$ by learning the parameters $w,b$ from few $(X,Y)$ pairs. Here for different tasks, the dimensionality $d_y$ of $Y$ changes from 5 to 20. Due to the randomness of regression sampling, we adopt the error reduction rate  defined as $ r_{err} = e_k / e_0 $, where $k \leq 5$ counts optimizing steps and $e_k$ denotes the error after k steps, to measure the adaptability and performance. To eliminate the effect of random initialization, we report the average error reduction rate over 100 different tasks. The resutls are given  in Table~\ref{table_regress}. As we can see, HML has a faster adaptation speed and better generalizability than MAML and fine-tuning baselines. 

\begin{table}[h]
\caption{Error reduction rate of multivariate linear regression tasks over different output dimensions, after 5 adaptation steps. Every task contains 5 context points $(X,Y)$ with $d_x=10$ and randomly sampled coefficient $w, b$.  The number in the bracket shows the 1 step adaptation performance.}
\centering
\small
\begin{tabular*}{0.46\textwidth}{@{\extracolsep{\fill}}ccccc}
\toprule
%  & \small 5d         & 8d         &   10d          &  20d \\ \midrule
% Rand   & 92.7(99.7)                  & 100.2(99.8)            & 92.21(99.8)          & 101.5(99.8) \\
% MAML        & 43.6(59.3)                    & 43.6(71.5)            & 100.2(99.8)        & 100.2(99.8) \\
% ours  & \textbf{18.9}(\textbf{29.3})   & \textbf{24.8}(\textbf{49.4}) & 60.4 & \textbf{41.8}  \\
 $d_y$ & \small 5              &   10          &  20 \\ \midrule
Fine-tune   & 92.7(99.7)             & 92.21(99.8)          & 101.5(99.8) \\
MAML        & 43.6(59.3)             & 46.5(77.1)        & 59.8(86.8) \\
HML (ours)  & \textbf{18.9}(\textbf{29.3})    & \textbf{25.3}(\textbf{55.7}) & \textbf{42.4}(\textbf{75.6})  \\
% \midrule
% SUN2012     &\small 5w1s & 10w1s & \small 5w5s &  10w5s \\ \midrule
% Fine-tune   & 32  & 14.0 & 41.2 & 24.2 \\
% MAML        & \textbf{49.9} & 32.5 & 64.6  & 39.8 \\
% HML (ours)  & 48.6  & \textbf{39.5}  & \textbf{ 65.4} & \textbf{45.2} \\
\bottomrule
\end{tabular*}%
\label{table_regress}
\end{table}

\subsection{Performance Analysis on HML}
To understand the effectiveness of HML intuitively, we conduct the following experiments on the Omniglot dataset. First, we visualize the data representation learned by the meta models from MAML and HML,  via t-SNE~\cite{maaten2008visualizing} in Figure~\ref{fig_result_maml},  for the $20$-way $1$-shot  tasks. Here the model is trained on the $5$-way $1$-shot source tasks.   Comparing with MAML, the representation learned by the meta model with HML presents higher intra-class compactness and inter-class separability. This implies the benefits of HML from explicitly optimizing the across-task generalizability. With hierarchical meta learning, the resulted meta model learns representations that are more broadly suitable for the tasks involving more categories to classify. As the representations of different categories are more separated, the model offers a stronger generalizability to other $N'$-way classification tasks with a different $N'$. 
%In another word, the difference between the distance of intra-class and inter-class is more distinct. 
%It illustrated that HML trained on known tasks can provide better representation for new tasks to be classified.  

\begin{figure}[t]
\centering
\includegraphics[height= 0.3\textwidth,width=0.45\textwidth]{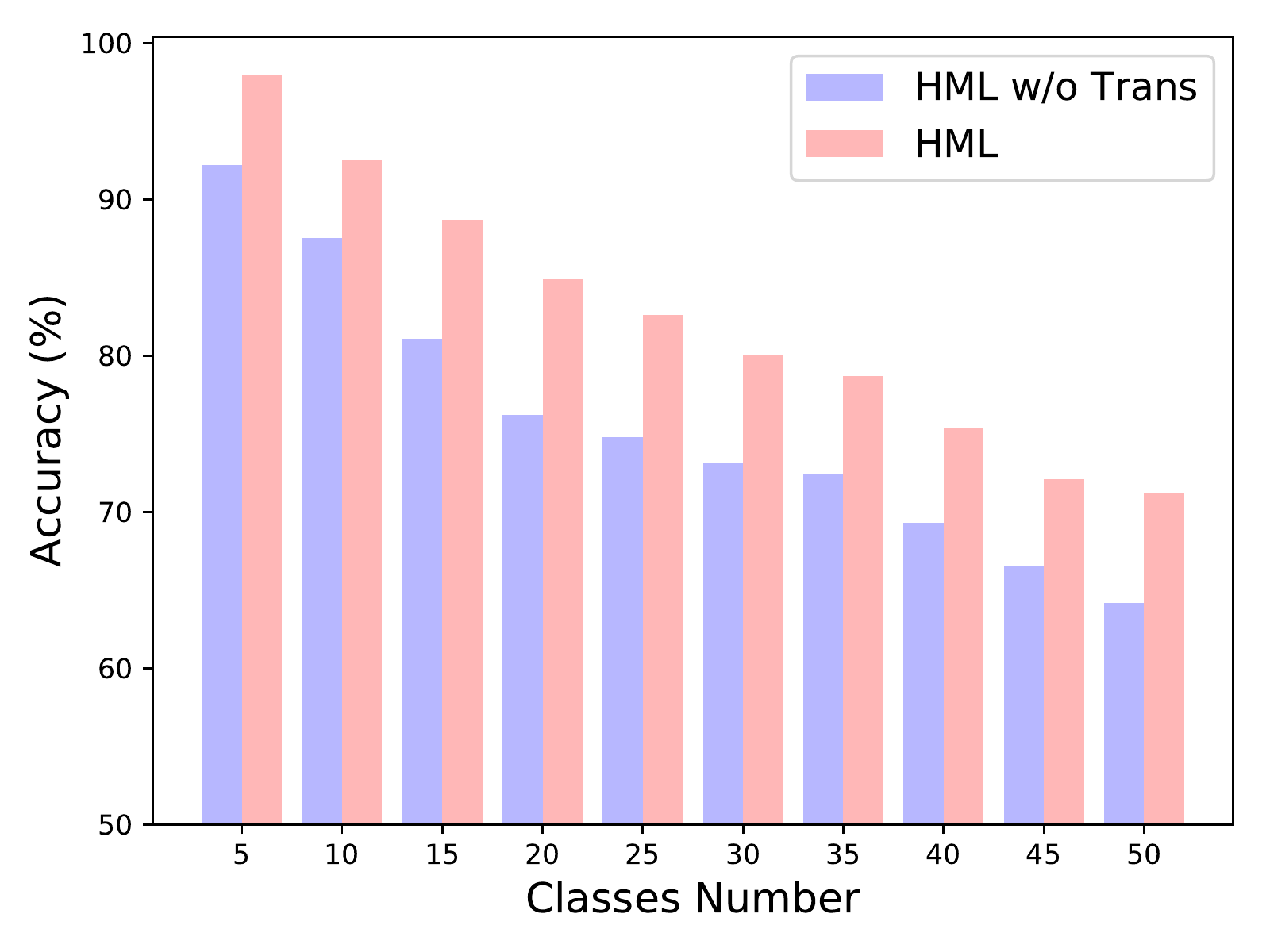}
\caption{Comparison on accuracy  of $n$-way $1$-shot classification  between HML and HML (w/o Trans) on heterogeneous tasks.}
\label{fig_acc_with_layer}
\end{figure}

Second, we study the effects of the transformation function $\omega$ within HML. We compare the performance of the models from vanilla HML and from HML without the function $\omega$, denoted as \emph{HML} and \emph{HML w/o Trans} respectively. We plot their performance for $1$-shot learning for different ways (ranging from $5$ to $50$) in  Figure~\ref{fig_acc_with_layer}. We can observe that  the accuracy  of HML is consistently slightly higher than HML w/o Trans. Furthermore, the margin between them becomes more significant when the number of categories is larger than $5$. Since  the transformation function does not change the  expressive capacity of the model internal representation, the performance improvement mainly comes from  the transformation function $\omega$ by   improving model adaptability to different task structures. Notably, even for HML w/o Trans, its generalization performance is much better than MAML. This demonstrates the meta model indeed benefits its generalization performance due to the hierarchical meta learning scheme of HML.

%The performance substantially approximates the optimal after three iterations which illustrates our well trained model is ready to fast adapt to new heterogeneous tasks.

\section{Conclusion}
In this work, we present a new meta learning  problem of making the meta models quickly solve new tasks of different structures from the training ones, and devise a new Hierarchical Meta Learning (HML) to explicitly optimize the generalization performance of a model across heterogeneous tasks.
%This problem setting aims at extending the existing meta learning approaches,  obtaining meta model of stronger generalization ability and  modeling  real application scenarios more faithfully.  The effective and efficient model can adapt to new task distributions swiftly without re-training.
%Additionally,  we also present a meta-learned transformation function    to help model adapt to new task structures. We proved our HML is able to learn generalization rule which is effective and efficient in  solving new tasks.  In the future, this generalization problem of few-shot learning needs to be improved and generalization itself needs to be solved.

%\newpage
 
% \clearpage

\bibliographystyle{named}
\bibliography{ijcai19}

% \newpage
% \appendix
% \section{Additional analysis}
% Here, we evaluate the adaption speed of our proposed HML in order to demonstrate that the meta model obtained from HML is also efficient in learning new tasks, in addition to gaining generalization ability. To this end, we fine-tune the meta model from HML on the training examples of new tasks with gradient descent and plot the classification accuracy of $5,20,50$-way 1-shot learning tasks with different numbers of gradient descent steps in Figure~\ref{fig:adap_speed}. The model with a single adaptation step has already achieved satisfactory performance and its performance reaches the optimum within 5 adaptation steps. This clearly shows the attractive efficiency of HML. It provides a meta model that is directly applicable to novel tasks without requiring to repeat the meta-training process. 
% \begin{figure}[h!]
% \center
% %\framebox(200,160){Adaptation speed to new tasks}
% \includegraphics[height= 0.29\textwidth, width=0.5\textwidth]{speed.pdf}
% \caption{Adaptation speed to new tasks of  HML. x-axis counts gradient descent steps to update the meta model parameters.  The results are from evaluation over $1$-shot $5,20,50$-way tasks on Omniglot with  10 random runs. }
% \label{fig:adap_speed}
% \end{figure}

\end{document}